\documentclass{article}
\usepackage{ijcai25}

\usepackage{times}
\usepackage{soul}
\usepackage{url}
\usepackage[hidelinks]{hyperref}
\usepackage[utf8]{inputenc}
\usepackage[small]{caption}
\usepackage{graphicx}
\usepackage{amsmath}
\usepackage{amsthm}
\usepackage{booktabs}
\usepackage{algorithm}
\usepackage{algorithmic}
\usepackage[switch]{lineno}
\usepackage{amssymb}

\title{Multi-Hierarchical Fine-Grained Feature Mapping Driven by Feature Contribution for Molecular Odor Prediction}

\author{
    HongXin Xie$^1$, JianDe Sun$^1$, FanFu Xue$^2$, ZiFei Han$^3$, ShanShan Feng$^1$, Qi Chen$^4$ \\
$^1$ Shandong Normal University, Jinan, China \\
$^2$ Shandong University, Jinan, China \\
$^3$ ZJU-UIUC Institute, Haining, China  \\
$^4$Huangshan University, Huangshan, China
}

\begin{document}

\maketitle

\begin{abstract}

Molecular odor prediction is the process of using a molecule's structure to predict its smell. While accurate prediction remains challenging, AI models can suggest potential odors. Existing methods, however, often rely on basic descriptors or handcrafted fingerprints, which lack expressive power and hinder effective learning. Furthermore, these methods suffer from severe class imbalance, limiting the training effectiveness of AI models. To address these challenges, we propose a Feature Contribution-driven Hierarchical Multi-Feature Mapping Network (HMFNet). Specifically, we introduce a fine-grained, Local Multi-Hierarchy Feature Extraction module (LMFE) that performs deep feature extraction at the atomic level, capturing detailed features crucial for odor prediction. To enhance the extraction of discriminative atomic features, we integrate a Harmonic Modulated Feature Mapping (HMFM). This module dynamically learns feature importance and frequency modulation, improving the model's capability to capture relevant patterns. Additionally, a Global Multi-Hierarchy Feature Extraction module (GMFE) is designed to learn global features from the molecular graph topology, enabling the model to fully leverage global information and enhance its discriminative power for odor prediction. To further mitigate the issue of class imbalance, we propose a Chemically-Informed Loss (CIL). Experimental results demonstrate that our approach significantly improves performance across various deep learning models, highlighting its potential to advance molecular structure representation and accelerate the development of AI-driven technologies.
\end{abstract}

\section{Introduction}

Odor, a key sensory characteristic, significantly influences consumer experience and product perception \cite{1}. By understanding the molecular structure-odor relationship, AI models can predict how molecules interact with the human olfactory system \cite{3,4}. In personalized medicine, AI-based odor prediction helps create custom scents for individual health and wellness needs \cite{5}. These innovations highlight AI’s transformative potential to advance critical technologies in biotechnology, health, and environmental sciences.

\begin{figure*}[htbp]
\centerline{\includegraphics[width=1\textwidth,height=0.3\textheight,keepaspectratio]{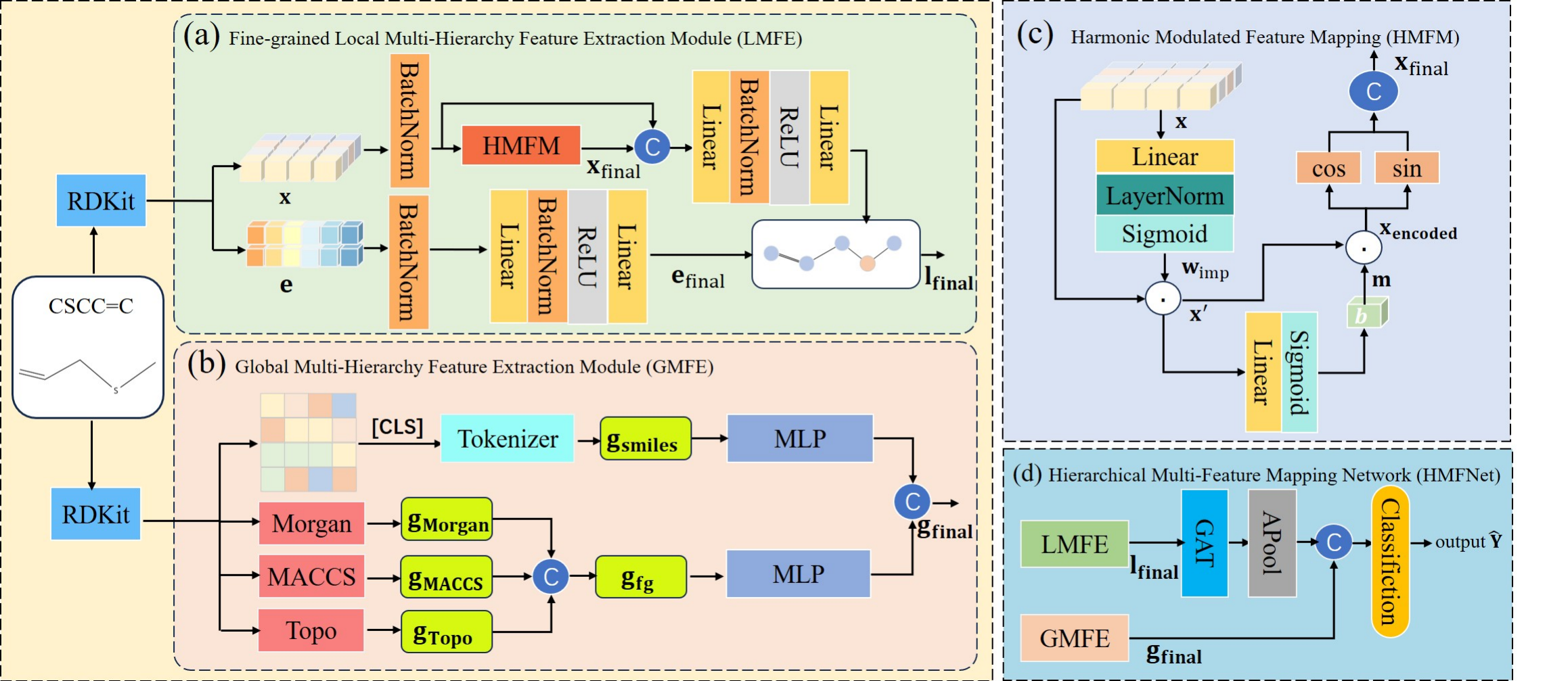}}
\caption{(a):  The concrete structure of fine-grained Local Multi-Hierarchical Feature Extraction; (b): The concrete structure of Global Multi-Hierarchy Feature Extraction; (c): The concrete structure of concrete structure of Harmonic Modulated Feature Mapping; (d): The architecture of the proposed overall framework.}
\label{fig}
\end{figure*}

 Early approaches primarily relied on chemistry-based statistical machine learning techniques \cite{MD}. However, with advancements in artificial intelligence, modern methods now predominantly utilize deep learning technologies, such as graph neural networks \cite{gnn,10}, to model the intricate interactions between molecular structures and odor. Existing methods face significant challenges in capturing the complex relationship between molecular structure and odor. Traditional atomic-level features and handcrafted fingerprints fail to adequately represent these interactions due to their limited expressiveness. Furthermore, class imbalance in odor descriptors exacerbates model bias \cite{6}, hindering effective prediction. 

To address the aforementioned issues, we propose a Hierarchical Multi-Feature Mapping Network (HMFNet). Specifically, it consists of a fine-grained Local Multi-Hierarchy Feature Extraction module (LMFE) and a Global Multi-Hierarchy Feature Extraction module (GMFE). LMFE performs deep feature extraction on single-structure matrices, such as atomic, bond, and molecular structure data, to capture fine-grained local features that are more beneficial for odor prediction. To better extract discriminative features from atomic information, we design a Harmonic Modulated Feature Mapping (HMFM), which enhances the model's efficiency in utilizing molecular features by dynamically learning feature importance and applying frequency modulation. This improves the model's ability to handle the complex relationships between molecules and odors. GMFE learns global features from the molecular graph's topological structure, molecular fingerprints, and global chemical properties, leveraging global information to improve discriminative power for odor prediction and further enhancing the model’s capability to manage complex molecule-odor relationships. Additionally, we integrate components such as structural similarity constraints and label correlation to design a Chemically-Informed Loss (CIL), specifically tailored to address the class imbalance problem in molecular odor prediction, thereby improving the model's performance. Our main contributions are as follows:
\begin{itemize}

\item We propose LMFE that performs deep feature extraction to capture fine-grained features critical for odor prediction. To further enhance the extraction of discriminative atomic features, we design HMFM. This module dynamically learns feature importance and applies frequency modulation, improving the model's ability to capture complex odor-molecular relationships.
\item We propose an improved loss function, Chemically-Informed Loss, which incorporates multiple components. This multi-faceted design addresses issues of class imbalance, improves the model’s focus on minority classes, and fosters better learning of label co-occurrence relationships.
\item We propose HMFNet, a multi-hierarchical framework that enhances odor prediction by integrating GMFE into LMFE, effectively capturing feature information from local to global levels. Experimental results confirm that, guided by the CIL, our approach achieves state-of-the-art performance.
\end{itemize}

\section{Related Work}

Molecular odor prediction has become a key research focus across chemistry, neuroscience, and computer science. With the advent of machine learning, many studies now employ computational methods to predict the olfactory properties of molecules. Early research explored the relationship between molecular structure and odor through chemical parameters. For example, PaDEL-Descriptor \cite{7} calculates 797 molecular descriptors and 10 types of fingerprints, including electro-topological state descriptors and molecular volume, which are essential for quantitative structure–activity relationship (QSAR) studies. Despite its extensive library, PaDEL-Descriptor faces challenges in processing speed and its ability to handle large molecules. To overcome these limitations, Mordred \cite{8} introduced a more advanced descriptor calculation tool, capable of computing over 1,800 2D and 3D molecular descriptors. Mordred is at least twice as fast as PaDEL and can handle large molecular descriptors that other tools cannot. With its high performance, ease of use, and comprehensive descriptor library, Mordred has become a key resource in cheminformatics, especially for structure–property relationship studies. While descriptor-based feature extraction remains important, machine learning techniques are increasingly driving advancements in molecular odor prediction. Graph neural networks (GNNs) \cite{gnn} have proven effective in modeling the complex relationship between molecular structure and odor perception. More recently, Lee et al. \cite{10} employed GNNs to create an odor mapping that preserves perceptual relationships, facilitating quality prediction for uncharacterized odor molecules. In prospective validation on 400 unseen odor samples, the POM model's odor profiles were closer to the training group's mean than the median, confirming its reliability as a prediction tool. This model outperformed traditional cheminformatics methods, effectively encoding the structure–odor relationship. Additionally, OWSum \cite{11} proposed the Odor Weighted Sum (OWSum) algorithm, a linear classifier combining structural patterns with conditional probabilities and TF-IDF values for odor prediction. This approach not only enhances our understanding of odor prediction but also advances molecular odor prediction.

\section{Methodology}

\subsection{Overview}

Molecular odor prediction aims to predict odor descriptors based on molecular structures, offering significant benefits in fragrance design, environmental monitoring, and personalized health products. However, existing methods face challenges in capturing the complex relationships between molecular structures and odor properties, as well as dealing with data imbalance, which affects model performance.

To address these issues, we propose a Hierarchical Multi-Feature Mapping Network (HMFNet), the detailed structure is shown in Figure 1. Our method includes a local to global molecular feature extraction technique, which enhances feature richness by incorporating atomic-level, bond-level, and global descriptors. This allows the model to better capture the complex relationships between molecular structure and odor, improving prediction accuracy. For a detailed description, refer to Sections 3.2 and 3.3. We also introduce a novel molecular feature mapping method, Harmonic Modulated Feature Mapping (HMFM), which dynamically adjusts feature contributions through importance learning and frequency modulation. This enables the model to effectively capture complex relationships, improving the overall predictive performance of molecular odor characteristics, the specifics are discussed in Section 3.2. Lastly, we design a Chemically-Informed Loss (CIL) function to address class imbalance and inter-label dependencies. By incorporating structural similarity constraints, label correlation integration, and adaptive energy adjustments, CIL enhances the model’s ability to handle imbalanced datasets, particularly improving predictions for minority odor descriptors, detailed description in section 3.4. 

\subsection{Fine-grained Local Multi-Hierarchy Feature Extraction Module (LMFE)}


 To capture fine-grained local features that are more beneficial for odor prediction, we design a fine-grained Local Multi-Hierarchy Feature Extraction Module (LMFE). LMFE performs deep feature extraction on single-structure matrices such as atomic and bond data. Specifically, based on RDKit \cite{rd}, for each atom \(v\), its feature representation is \(\mathbf{x}_{v}^{A}\), where \(A\) represents the characteristic number of atoms. For each chemical bond \(b\), its feature representation is \(\mathbf{e}_{b}^{I}\), where \(I\) represents the characteristic number of chemical bonds. 
 
 In molecular odor prediction tasks, existing methods \cite{6,11} struggle to learn non-smooth objective functions and address the issue of mixed feature dimensions, make traditional feature mapping methods insufficient for effectively capturing these multidimensional relationships. To overcome this issue, we propose a novel Harmonic Modulated Feature Mapping (HMFM) method based on feature importance learning and frequency modulation. To achieve this, we introduce a feature importance layer and a frequency modulation layer. The modulation, combined with base frequencies, forms periodic and phase encoding, effectively capturing the complex relationships between molecular features and odors.

Specifically, for each atomic feature, we learn its relative importance in odor prediction through the feature importance layer, enabling the model to adaptively adjust each feature's impact on the final prediction. Given the input feature matrix \(\mathbf{x}\in\mathbb{R}^{N\times A}\), where \(N\) is the batch size and \(A\) is the number of atomic features, the feature importance weight \(\mathbf{w}_{\mathrm{imp}}\in\mathbb{R}^{N\times A}\) is calculated through the following steps:
\begin{equation}
    \mathbf{w}_{\mathrm{imp}}=\sigma\left(\mathrm{LayerNorm}\left(\mathrm{Linear}(\mathbf{x})\right)\right)
\end{equation}
where \({{\sigma}}\) is the Sigmoid activation function, \(\mathrm{Linear}(\cdot)\) is  a linear transformation, and \(\mathrm{LayerNorm}(\cdot)\) is layer normalization. The resulting \(\mathbf{w}_{\mathrm{imp}}\) represents the learned importance of each atomic feature.

Next, the weighted features \(\mathbf{x}^{\prime}\) are obtained by element-wise multiplication of the feature matrix \(\mathbf{x}\) and the importance weights \(\mathbf{w}_{\mathrm{imp}}\):
\begin{equation}
    \mathbf{x}^{\prime}=\mathbf{x}\odot\mathbf{w}_{\mathrm{imp}}
\end{equation}
where \(\odot\) denotes the element-wise multiplication.

To apply different frequency responses to various features, we design a frequency modulation mechanism. By learning to modulate atomic features, we dynamically adjust the frequency of each feature, enabling dynamic adaptation between feature importance and frequency. Specifically, we apply the frequency modulation layer on the input features \(\mathbf{x}^{\prime}\). This layer generates a modulation coefficient \(\mathbf{f}\in \mathbb{R}^{N\times D}\), where \(D\) represents the output feature dimension. The frequency modulation coefficient is computed as:
\begin{equation}
    \mathbf{f}=\sigma\left(\mathrm{Linear}(\mathbf{x}^{\prime})\right)
\end{equation}

The obtained frequency modulation coefficient \(\mathbf{f}\) is then multiplied element-wise with a base frequency coefficient \(\mathbf{b}\) to obtain the modulated frequency coefficients \(\mathbf{m}\):
\begin{equation}
    \mathbf{m}=\mathbf{b}\odot\mathbf{f}
\end{equation}

The base frequency coefficient \(\mathbf{b}\) is pre-calculated using the formula \(\mathbf{b}=2\pi\sigma^{\prime}\frac{j}{D}\), where \(\sigma^{\prime}\) represents the standard deviation, and \(j\) denotes the index of the feature dimension.

Once the modulated frequency coefficients are obtained, we combine them with the weighted feature \(\mathbf{x}^{\prime}\) to generate periodic and phase encodings. Specifically, the encoding result is calculated through the following steps:
\begin{equation}
    \mathbf{x}_{\mathrm{encoded}}=\mathbf{m}\odot\mathbf{x}^{\prime}
\end{equation}

Then, the cosine and sine values of \(\mathbf{x}_{\mathrm{encoded}}\) are computed: \(\cos(\mathbf{x}_{\mathrm{encoded}}),  
 \sin(\mathbf{x}_{\mathrm{encoded}})\).

Finally, the feature mapping \(\mathbf{x}_{\mathrm{final}}\) is obtained by concatenating the cosine and sine values along the feature dimension:
\begin{equation}
    \mathbf{x}_{\mathrm{final}}=\mathrm{concat}(\cos(\mathbf{x}_{\mathrm{encoded}}),\sin(\mathbf{x}_{\mathrm{encoded}}),\dim=-1)
\end{equation}
where \(\mathrm{concat}\) denotes concatenation along the last dimension.

The final features, \(\mathbf{x}_{\mathrm{final}}\) and the bond features \(\mathbf{e}\), are passed through two encoders to form the molecular graph representation \(\mathbf{l}_{\mathrm{final}}\). This representation is then processed through a GAT Network \cite{21} and a pooling layer to obtain the enhanced features \(\mathbf{l}_\mathbf{final}^{\prime}\).

\subsection{Global Multi-Hierarchy Feature Extraction Module (GMFE)}

Additionally, to further enhance the model's ability to handle complex molecule-odor relationships, we design a Global Multi-Hierarchy Feature Extraction Module (GMFE). GMFE learns global features from the molecular fingerprints and global chemical properties, effectively leveraging global information to improve the model's discriminative power for odor prediction. We compute three types of molecular fingerprints based on RDKit: Morgan fingerprint \cite{mo}, MACCS fingerprint \cite{ma}, and Topological fingerprint \cite{to}. The feature vectors for Morgan, MACCS, and Topological fingerprints are denoted as \(\mathbf{g_{\mathrm{Morgan}}}\), \(\mathbf{g_{\mathrm{MACCS}}}\), \(\mathbf{g_{\mathrm{Topo}}}\). By concatenating these fingerprint vectors, we obtain the overall molecular fingerprint representation:
\begin{equation}
    \mathbf{g}_{\mathrm{fg}}=\mathrm{concat}(\mathbf{g}_{\mathrm{Morgan}},\mathbf{g}_{\mathrm{MACCS}},\mathbf{g}_{\mathrm{Topo}})
\end{equation}
where \(\mathrm{concat}\) denotes the concatenation operation.

We utilize a Transformer-based \cite{t} approach for the SMILES strings to obtain a global chemical property representation of the molecule. The molecules are labeled and embedded into vectors, with the resulting feature representation denoted as \(\mathbf{g}_{\mathrm{smiles}}\).

The resulting \(\mathbf{g}_{\mathrm{fg}}\) and \(\mathbf{g}_{\mathrm{smiles}}\) are passed through two identical MLP layers to obtain the final \(\mathbf{g}_{\mathrm{final}}\) is then combined with the enhanced \(\mathbf{l}_\mathbf{final}^{\prime}\) to form the final representation.

\begin{figure}
\centerline{\includegraphics[width=0.47\textwidth,height=0.47\textheight,keepaspectratio]{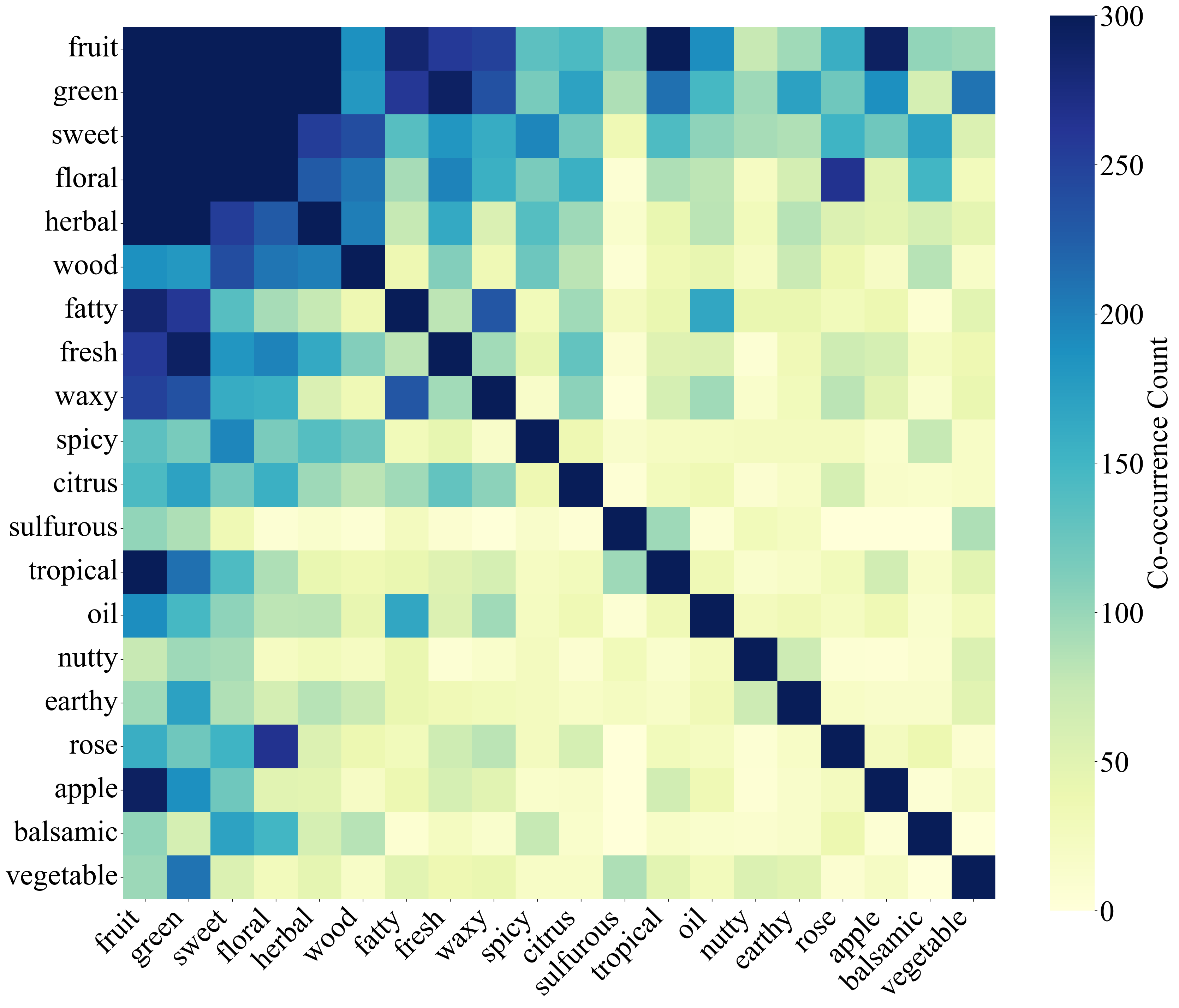}}
\caption{Co-ocurrence matrix of Top 20 odor descriptors.}
\label{fig}
\end{figure}

\subsection{Chemically-Informed Loss (CIL)}

To address data imbalance in molecular odor prediction, we propose a Chemically-Informed Loss function comprising five key components: weighted binary cross-entropy (BCE) loss \cite{14}, molecular structural similarity loss, chemical property energy loss, sample-level multi-label constraint loss, and label correlation loss. These components are designed to overcome the limitations of existing methods in handling multi-label prediction and complex chemical information. Each component is designed based on a deep understanding of molecular features, aiming to overcome the limitations of existing methods in handling complex chemical information and multi-label prediction.
To address the class imbalance in odor descriptors, we introduce a weighted BCE loss. Specifically, the weight \(w_{j}\) for each class is computed based on the ratio of positive to negative samples and is adjusted dynamically during training:
\begin{equation}
    w_{j}=\frac{W_{\mathrm{neg},j}}{W_{\mathrm{pos},j}},\quad w_{j}\in[0.1,10]
\end{equation}
where \({W_{\mathrm{pos},j}}\) and \({W_{\mathrm{neg},j}}\) represent the numbers of positive and negative samples for the odor descriptor \(j\). To prevent training instability caused by very large or very small weights, the weights are limited to the range \([0.1, 10]\).
\begin{small}
\begin{equation}
    \mathcal{L}_{\text{basis}}=-\frac{1}{N}\sum_{i=1}^{N}\sum_{j=1}^{M}w_{j}\left[Y_{i,j}\log(\hat{Y}_{i,j})+(1-Y_{i,j})\log(1-\hat{Y}_{i,j})\right]
\end{equation}
\end{small}
where \(N\) is the batch size, and \(M\) is the number of classes (i.e., the number of odor descriptors), \(Y_{i,j}\in\{0,1\}\) indicates whether molecule \(i\) has the odor descriptor \(j\) (\(1\) means present, and \(0\) means absent). \(\hat{Y}_{i,j}\) represents the predicted probability of the model, which indicates the likelihood that sample \(i\) has the odor descriptor \(j\).

Additionally, we introduce an ``energy'' \cite{15} function related to molecular odor features, which sets a target energy for each odor descriptor, constraining the model's learning process to ensure its predictions align with chemical properties and physical laws. Specifically, if the predicted probability of molecule \(i\) for odor descriptor \(j\) is denoted as \(\hat{Y}_{i,j}\), the energy \(E_{energy}(j)\) of odor descriptor \(j\) is defined as the average prediction probability of this descriptor across the entire sample set:
\begin{equation}
    E_{energy}(j)=\frac{1}{N}\sum_{i=1}^N\hat{Y}_{i,j}
\end{equation}

\begin{table*}
    \centering
    \fontsize{6pt}{7pt}\selectfont
\resizebox{0.7\textwidth}{!}{
    \begin{tabular}{cccccccc}
\hline
\multicolumn{6}{c}{Method}        & \multicolumn{2}{c}{Evaluation Metrics} \\ \hline
Node        & Edge       & Fingerprint & Token          & HMFM          & CIL           & F1 score            & AUROC     \\ \hline
\checkmark  & ×          & ×           & ×              & ×             & ×             & 0.3400              & 0.9239    \\
\checkmark  & \checkmark & ×           & ×              & ×             & ×             & 0.4167              & \textbf{0.9356} \\
\checkmark  & \checkmark & ×           & ×              & \checkmark    & ×             & \underline{0.4393}  & \underline{0.9337}   \\
\checkmark  & \checkmark & ×           & ×              & \checkmark    & \checkmark    & \textbf{0.4757}     & 0.9233   \\ \hline
\checkmark  & \checkmark & \checkmark  & ×              & ×             & ×             & 0.4385              & 0.9207    \\
\checkmark  & \checkmark & \checkmark  & \checkmark     & ×             & ×             & 0.4400              & 0.9221    \\
\checkmark  & \checkmark & \checkmark  & \checkmark     & \checkmark    & ×             & \underline{0.4508}  &\underline{0.9266}    \\
\checkmark  & \checkmark & \checkmark  & \checkmark     & \checkmark    & \checkmark    & \textbf{0.4861}     & \textbf{0.9316}   \\ \hline
    \end{tabular}}
    \caption{Ablation study results of key components. The best performance is highlighted in bold and the follow-up is highlighted in underlined.}
    \label{tab:booktabs}
\end{table*}

We introduce a constraint loss based on chemical property energy, with energy  \(m_{\mathrm{in}}\) and \(m_{\mathrm{out}}\). The target energy \(m_{\mathrm{in}}\) is set to 1 for samples with odor descriptors, and \(m_{\mathrm{out}}\) is set to 0 for those without. These targets are optimized using a label co-occurrence matrix \(C_{\mathrm{co-occurrence}}\), reflecting the frequency of odor descriptor co-occurrence. Descriptors that frequently co-occur are assigned higher energy targets, improving the model's understanding of their interrelationships. The energy target formula is:
\begin{equation}
    m_{\mathrm{in}}=1+c\cdot\mathrm{diag}\left(\frac{1}{N}\sum_{i=1}^{N}Y_{i}^{T}Y_{i}\right)
\end{equation}

\begin{equation}
    m_{\mathrm{out}}=c\cdot\mathrm{diag}\left(\frac{1}{N}\sum_{i=1}^N(1-Y_i)^T(1-Y_i)\right)
\end{equation}
Here, we conducted comparative experiments on the hyperparameters \(c=0.2\) with detailed results provided in Appendix \ref{appendix:Supplementary Experiments}. \(c\) controls the extent to which label co-occurrence relationships influence the adjustment of energy targets. This ensures that the energy target adjustment is neither excessively amplified (avoiding excessively high energy targets) nor too small (which would weaken the impact of energy adjustment on model training), \(Y_i\) represents the label vector for molecule \(i\), and \(\mathrm{diag}(\cdot)\) denotes the extraction of diagonal elements from the matrix, which indicates the co-occurrence frequency of different descriptors.


The chemical property energy loss is:

\begin{equation}
\mathcal{L}_{\mathrm{class}} = 
\begin{aligned}
    & \sum_{j=1}^M \left[ \sum_{i:Y_{i,j}=1} \max(0, E_{energy}(j) - m_{\mathrm{in}})^2 \right] \\
    & + \sum_{j=1}^M \left[ \sum_{i:Y_{i,j}=0} \max(0, m_{\mathrm{out}} - E_{energy}(j))^2 \right]
\end{aligned}
\end{equation}

To further enhance multi-label prediction performance, we designed a sample-level multi-label constraint loss. The expected energy for a sample is adjusted based on the label count:
\begin{equation}
    E_{\mathrm{expected}}(i)=e_{1}+e_{2}\cdot\sum_{j=1}^MY_{i,j}
\end{equation}
Here, \(e_{1}+e_{2}=1\) are hyperparameters. \(e_{1}\) represents the baseline expected energy for each sample, ensuring that the model does not generate extreme energy targets due to an insufficient or excessive number of labels, thereby enhancing training stability. \(e_{2}\) is a modulation factor that ensures the increase in label count smoothly influences the expected energy of the sample, preventing an excessive number of labels from leading to overly large expected energies.

\begin{table}
\fontsize{4pt}{5pt}\selectfont
    \centering
    \resizebox{0.47\textwidth}{!}{
    \begin{tabular}{ccr}
    \hline
Method           & F1 score       & AUROC   \\ \hline
GCN \cite{20}    & 0.3701         & 0.9271     \\
GCN+HMFM         & 0.3910         & 0.9296    \\ \hline
GAT \cite{21}    & 0.3953         & 0.9274     \\
GAT+HMFM         & 0.4066         & 0.9296    \\ \hline
MPNN \cite{22}   & 0.4235         & 0.9304     \\
MPNN+HMFM        & 0.4338         & 0.9314  \\ \hline
AFP \cite{afp}   & 0.4429         & 0.9240   \\
AFP+HMFM         & 0.4728         & 0.9255   \\ \hline
SMPGNN \cite{SMP}& 0.4448         & 0.9175\\ 
SMPGNN+HMFM      & 0.4550         & 0.9231\\ \hline
GCast \cite{gc}  & 0.4622         & 0.9288  \\
GCast+HMFM       & 0.4677         & 0.9289  \\ \hline
GSAGE \cite{SAGE}& 0.3858         & 0.9284  \\
GSAGE+HMFM       & 0.4187         & 0.9295   \\ \hline
GIN \cite{GIN}   & 0.3973         & 0.9303  \\
GIN+HMFM         & 0.4158         & 0.9304  \\ \hline
HMFNet           &\textbf{0.4861} &\textbf{0.9316}  \\ \hline
    \end{tabular}}
    \caption{Performance comparison of harmonic modulated feature mapping in mainstream deep learning models.}
    \label{tab:booktabs}
\end{table}

The loss is calculated based on the difference between the sample's predicted energy and the expected energy:
\begin{equation}
    \mathcal{L}_{\mathrm{sample}}=\frac{1}{N}\sum_{i=1}^N\left[\max(0,E_{\mathrm{expected}}(i)-\sum_{j=1}^M\hat{Y}_{i,j})^2\right]
\end{equation}

We introduce the label correlation loss, designed to minimize the discrepancy between the predicted correlation and the true label correlation. The correlation between labels is measured using the inner product of the label matrix, while the predicted correlation is computed through the inner product of the predicted outputs:
\begin{equation}
    \mathcal{L}_{\text{col}}=\|\frac{1}{N}\sum_{i=1}^N\hat{Y}_i\hat{Y}_i^T-\frac{1}{N}\sum_{i=1}^NY_iY_i^T\|_2^2
\end{equation}
Here, \(\hat{Y}_{i}\) represents the predicted value for the \(i\)-th molecule, \({Y}_{i}\) represents the true value for the \(i\)-th molecule, \(\|\cdot\|_2^2\) represents the squared L2 norm, which is equivalent to the squared Euclidean distance.

Finally, the weighted sum of all loss terms constitutes the total loss function:
\begin{equation}
    \mathcal{L}_{\mathrm{total}}=\lambda_1\mathcal{L}_{\text{basis}}+\lambda_2\mathcal{L}_{\mathrm{class}}+\lambda_3\mathcal{L}_{\mathrm{sample}}+\lambda_4\mathcal{L}_{\text{col}}
\end{equation}
Here, we conducted comparative experiments on the hyperparameters \(\lambda_1\), \(\lambda_2\), \(\lambda_3\), \(\lambda_4\) with detailed results provided in Appendix \ref{appendix:Supplementary Experiments}. 


\section{Experiments}

In this section, we solve several key challenges in molecular odor prediction by exploring the following research questions:
\begin{itemize}
\item Q1: Can multi-level feature extraction effectively improve the performance of molecular odor prediction?
\item Q2: Does Harmonic Modulated Feature Mapping improve the performance of representative deep models?
\item Q3: Does Chemically-Informed Loss alleviate the impact of the imbalance of molecular odor prediction datasets?
\item Q4: Does the proposed design achieve the best performance of molecular odor prediction at present?
\end{itemize}

\subsection{Experiment Setting}
\noindent\textbf{Datasets.} The dataset used in this study is sourced from the Leffingwell PMP 2001 \cite{16} and the GoodScents \cite{17}, both of which provide valuable data for exploring the relationship between molecular structures and odor descriptors. The detailed dataset distribution can be found in Appendix \ref{appendix:datasets}, the dataset is characterized by significant label imbalance, as evidenced by the long-tail distribution of odor descriptors. Moreover, previous studies emphasize the importance of considering label dependency information\cite{19}. Given that many odor descriptors occur infrequently, we present a co-occurrence matrix for the top 20 descriptors in Figure 2 for visual clarity.

\begin{table}
    \centering
    \fontsize{4pt}{5pt}\selectfont
\resizebox{0.47\textwidth}{!}{
    \begin{tabular}{crr}
    \hline
Method               & F1 score        & AUROC               \\ \hline
GCN+GRFF \cite{23}    & 0.3905          & 0.9265                   \\
GCN+RFF \cite{24}     & 0.3833          & 0.9275                \\
GCN+PE \cite{25}      & 0.3807          & 0.9271                 \\
GCN+LEE \cite{26}     & 0.3908          & 0.9295                  \\
GCN+HMFM             & 0.3910          & \textbf{0.9296}    \\ 
GCN+HMFM+CIL         &\textbf{0.4560}  & 0.9248                  \\ \hline
MPNN+GRFF \cite{23}   & 0.4030          & 0.9308                   \\
MPNN+RFF \cite{24}    & 0.4114          &\textbf{0.9322}                \\
MPNN+PE \cite{25}     & 0.4181          & 0.9309                 \\
MPNN+LEE \cite{26}    & 0.4238          & 0.9299                \\
MPNN+HMFM            & 0.4338          & 0.9314   \\ 
MPNN+HMFM+CIL        &\textbf{0.4791}  & 0.9306           \\ \hline
GAT+GRFF \cite{23}   & 0.4022         & 0.9246                   \\
GAT+RFF \cite{24}    & 0.3655          & 0.9251                  \\
GAT+PE \cite{25}     & 0.4023         & 0.9264                \\
GAT+LEE \cite{26}    & 0.3887         & 0.9254               \\
GAT+HMFM            & 0.4066          & 0.9296   \\ 
GAT+HMFM+CIL        &\textbf{0.4692}  & \textbf{0.9289}          \\ \hline
    \end{tabular}}
    \caption{Experimental results of harmonic modulated feature mapping and Chemically-Informed Loss.}
    \label{tab:booktabs}
\end{table}

\noindent\textbf{Comparsion Setup.} To validate the effectiveness of each component in our multi-level feature extraction, we conducted ablation experiments on the atomic features, chemical bond features, fingerprint features, and Transformer-based string features. Additionally, we selected eight molecular prediction models as baseline models to evaluate the effectiveness of HMFM. These models include Graph Convolutional Network (GCN) \cite{20}, Graph Attention Network (GAT) \cite{21}, Attentive FP (AFP) \cite{afp}, Substructure Matching Pretrained GNN (SMPGNN) \cite{SMP}, Cross-scale Graph Propagation (GCast) \cite{gc}, Graph Sample and Aggregation (GSAGE) \cite{SAGE}, Graph Isomorphism Network (GIN) \cite{GIN}, and Message Passing Neural Network (MPNN) \cite{22}. To further demonstrate the superiority of HMFM, we compared it with four state-of-the-art feature mapping methods: Gaussian Random Fourier Features (GRFF) \cite{23}, Random Fourier Features (RFF) \cite{24}, Positional Encoding (PE) \cite{25}, and Laplacian Eigenvector Encoding (LEE) \cite{26}. Finally, we performed a study on the Chemically-Informed Loss (CIL) to prove its efficacy. By comparing our framework with existing methods, we demonstrated that our approach achieves the best performance in molecular odor prediction.

\noindent\textbf{Evaluation Metrics.} We selected two indicators widely used in multi-label molecular odor prediction tasks, including F1 score and Area Under the Receiver Operating Characteristic Curve (AUROC). F1 score emphasizes the balance between precision and recall, which means it increases when the model becomes better at correctly identifying positive samples (i.e., reducing false positives and false negatives). AUROC, measures the model's overall ability to distinguish between the positive and negative classes.

\subsection{Q1: Ablation Study of Hierarchical Feature Extraction}

In this section, we demonstrate the effectiveness of the Hierarchical Feature Extraction through the ablation experiment of feature extraction. As shown in Table 1, when using only one feature type, the model's performance is limited. Specifically, relying solely on graph-based features achieves the highest AUROC but falls short in capturing the full complexity of odor prediction, as it lacks complementary information from molecular fingerprints and token embeddings. By combining graph-based features, molecular fingerprints, and token embeddings, this approach captures a wider range of molecular characteristics. Graph-based features provide structural context, fingerprints capture specific substructural elements, and token embeddings encode sequential relationships from the SMILES representation. This multi-layered feature extraction enables the model to better handle the complexity of odor prediction, enhancing its ability to generalize across diverse molecular structures.

\begin{table}
\fontsize{4pt}{5pt}\selectfont
    \centering
    \resizebox{0.47\textwidth}{!}{
    \begin{tabular}{ccr}
    \hline
Method                   & F1 score        & AUROC   \\ \hline
HMFNet+HIL \cite{hil}    & 0.3308          & 0.9174  \\
HMFNet+MTL \cite{mtl}    & 0.3784          & 0.9113 \\
HMFNet+BCE \cite{BCE}    & 0.3845          & 0.9292\\
HMFNet+ASL \cite{asl}    & 0.4297          & 0.9209 \\
HMFNet+AFL \cite{afl}    & 0.4632          & 0.9294 \\
HMFNet+CIL               & \textbf{0.4861} & \textbf{ 0.9316}    \\  \hline
    \end{tabular}}
    \caption{Experimental results of the Hierarchical Multi-Feature Mapping Network with different loss functions.}
    \label{tab:booktabs}
\end{table}

\subsection{Q2: Analysis of Harmonic Modulated Feature Mapping}

In this section, we aim to verify the effect of Harmonic Modulated Feature Mapping. We conducted experiments on representative deep learning models. As shown in Table 2, the integration of HMFM into the baseline architectures resulted in substantial improvements. To provide a clearer comparison, we present bar charts of the F1 and AUROC scores, shown in Figure 3 and Figure 4, respectively. Secondly, we compared it with several established feature mapping methods, as shown in Table 3.

The inclusion of HMFM consistently enhanced performance across all evaluation metrics, with a particularly notable increase in the F1 score. HMFM enhances the model’s ability to better leverage molecular features through two key mechanisms: (1) feature importance learning, which enables the model to dynamically prioritize the most relevant features, and (2) frequency modulation, which adjusts the frequency response of each feature, improving the encoding of molecular structure information. HMFM shows great promise in advancing molecular structure representation and improving molecular odor prediction tasks in chemoinformatics.

\begin{figure}
\centerline{\includegraphics[width=0.52\textwidth,height=0.52\textheight,keepaspectratio]{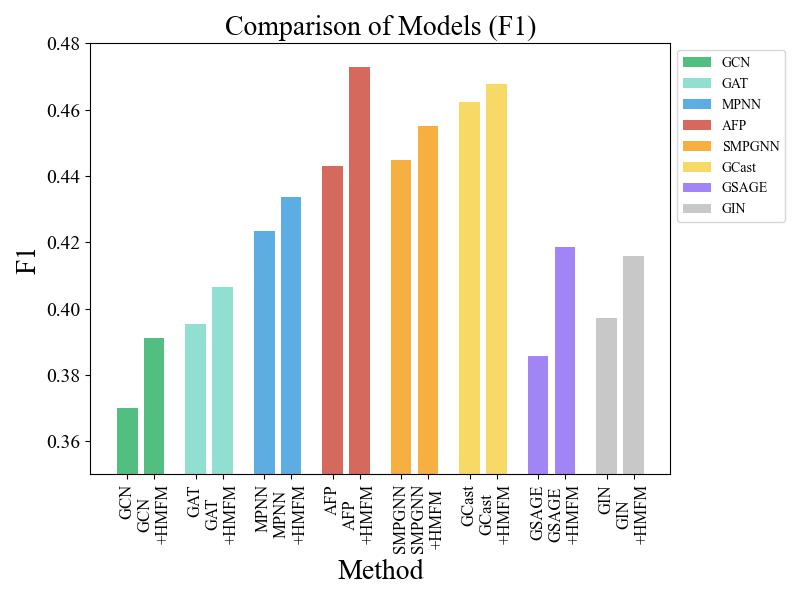}}
\caption{Comparison of F1 scores of histogram of Harmonic Modulated Feature Mapping on mainstream deep learning model.}
\label{fig}
\end{figure}

\subsection{Q3: Analysis of Chemically-Informed Loss}

To evaluate the effectiveness of the proposed Chemically-Informed Loss (CIL), we compared it with present loss function, including Hierarchical Loss (HIL) \cite{hil}, Binary Cross-Entropy Loss (BCE) \cite{BCE}, Asymmetric Loss (ASL) \cite{asl}, MultiTask Loss (MTL) \cite{mtl}, and Adaptive Focal Loss (AFL) \cite{afl}. As shown in Table 3 and Table 4, the integration of CIL significantly enhances model performance. Designed to address key challenges in molecular odor prediction—such as class imbalance, structural consistency, and label correlation—CIL demonstrates its robustness in improving predictive outcomes.

Experimental results show that incorporating CIL consistently increases the F1 score across various base models. By embedding chemical information constraints, CIL not only enhances classification accuracy but also strengthens the model's ability to capture the nuanced relationships between molecular structures and odor descriptors. Furthermore, it achieves a balanced prediction for both majority and minority classes. 

\begin{figure}
\centerline{\includegraphics[width=0.52\textwidth,height=0.52\textheight,keepaspectratio]{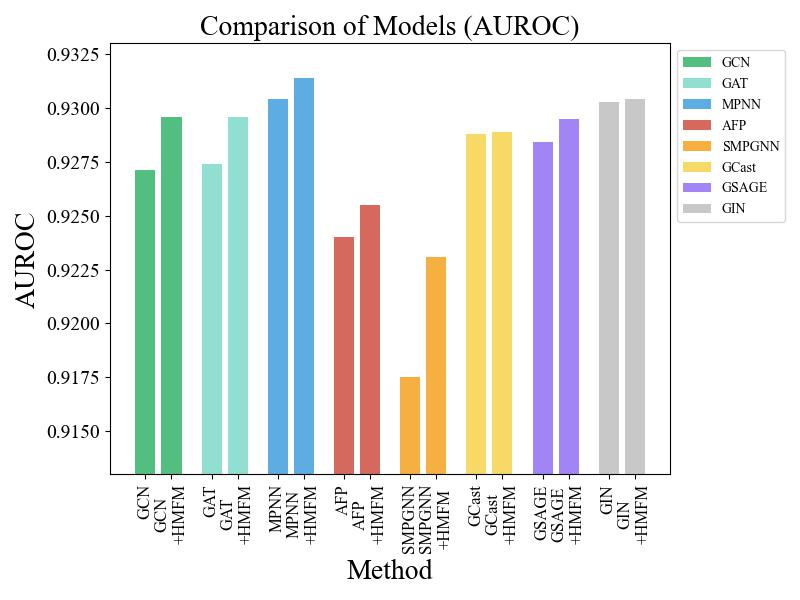}}
\caption{Comparison of histogram AUROC of Harmonic Modulated Feature Mapping on mainstream deep learning model.}
\label{fig}
\end{figure}

\subsection{Q4: Overall Comparison}

As shown in Table 1, The increase in F1 score with the addition of features and modules indicates that the model is improving its performance. Our method demonstrates a significant improvement in F1 score and achieves the best results in molecular odor prediction. However, the highest AUROC (0.9356) was achieved when only node and edge features were used. This suggests that the model performs better at distinguishing between positive and negative classes when fewer features are involved. Based on the calculation principles of F1 score and AUROC (as detailed in Section 4.1), we analyze that AUROC primarily focuses on the model's ability to correctly rank samples. The inclusion of additional features (fingerprints and tokens) may have introduced noise or led to less confident distinctions between classes, which slightly reduced the model's ability to differentiate samples based on predicted probabilities. However, the continuous improvement in F1 score indicates that the model becomes better at correctly identifying positive samples, reflecting a stronger ability to discern odor-related molecular structures. This improvement comes at the cost of a slight reduction in its ability to rank samples effectively, which in turn impacts AUROC. Detailed ablation studies on specific components of HMFNet can be found in Appendix \ref{appendix:Supplementary Experiments}.

By incorporating hierarchical feature extraction, our method integrates multiple complementary information sources, enabling the model to better capture the complex interactions that govern odor perception. Additionally, the novel feature mapping technique and optimized loss function facilitate dynamic adjustments to feature contributions and label consistency, addressing issues of data imbalance and enhancing prediction accuracy.


\section{Conclusion}

In this paper, we propose a novel framework for molecular odor prediction that effectively captures the complex relationships between molecules and their associated odors. By integrating local-to-global feature extraction with dynamic feature mapping and optimizing the loss function, our approach enhances the model’s ability to handle data imbalance and improve label correlations. This leads to a stronger capacity for identifying odor-related molecular structures. Experimental results demonstrate the superior performance and robust transferability of our method.

\bibliographystyle{named}
\bibliography{ijcai25}

\clearpage
\newpage
\appendix
\section{Appendix}
\subsection{Datasets}
\label{appendix:datasets}

\noindent\textbf{Leffingwell PMP 2001.} The Leffingwell PMP 2001 is a widely used resource in the study of molecular odor descriptors. It consists of a collection of chemical compounds along with their corresponding odor descriptors, which are used to capture the sensory characteristics of various molecules. The dataset includes odor descriptions for over 2,000 different chemical substances, each annotated with multiple odor descriptors. These descriptors reflect the perceptions of professional odorists and are typically multi-label, meaning that a single compound can be associated with multiple odor attributes (e.g., ``fruity", ``floral", ``woody", etc.).

This dataset is particularly valuable because it provides a broad spectrum of odor characteristics, enabling the modeling of complex relationships between molecular structure and odor. However, as is common in sensory data, there are imbalances in the frequency of occurrence for different odor descriptors. For example, more common descriptors such as ``fruity" or ``sweet" are associated with many more compounds than rarer descriptors like ``earthy" or ``musky".

\noindent\textbf{GoodScents.} The GoodScents, often referenced in the context of fragrance and flavor chemistry, is another significant resource in the field. It contains a detailed catalog of chemicals used in fragrances and flavors, along with their associated odor characteristics. This dataset provides a broad range of odor descriptors, categorized according to various sensory attributes like floral, herbal, spicy, and more.

GoodScents is particularly known for its extensive use in the perfume and flavor industry, where it aids in the formulation and understanding of scent profiles. Like the Leffingwell PMP 2001, GoodScents also exhibits label imbalance, with certain odor descriptors appearing much more frequently than others. The collection of odor descriptors is broad, covering a wide variety of sensory attributes, and provides a richer variety of data for studying molecular-odor relationships.

When combining the Leffingwell PMP 2001 and GoodScents, we observe a more comprehensive dataset that contains a diverse range of molecular odor descriptors, as shown in Figure 5. The combined dataset benefits from the strengths of both datasets, as it includes a large number of chemicals with detailed multi-label odor descriptions. This is ideal for exploring the complex relationships between molecular structure and odor perception.

However, as highlighted by the histogram of descriptor frequencies, the combined dataset exhibits a long-tail distribution of odor descriptors. Some descriptors like ``fruit", ``green", and ``floral" are highly frequent, appearing with a substantial number of compounds, while others such as ``musky", ``earthy", and ``camphor" are very rare. This imbalance, which is common in real-world datasets, poses challenges for model training. Frequent descriptors dominate the dataset, which can lead to bias in model predictions, making it harder to correctly predict rare odor descriptors.

Additionally, as shown in Figure 6, the co-occurrence matrix further reveals that certain odor descriptors frequently appear together, highlighting relationships that can be leveraged by the model. For example, ``fruit" and ``green" co-occur 774 times, and ``floral" and ``sweet" co-occur 504 times. These patterns of co-occurrence suggest that some odors are perceived together more often in the real world, and understanding these relationships is crucial for building better predictive models.

The long-tail distribution and co-occurrence patterns in the combined dataset have important implications for modeling. The label imbalance means that prediction models may need to be specifically tailored to handle underrepresented odor descriptors. Techniques such as weighted loss functions, data augmentation, or sampling strategies could be crucial in ensuring the model can generalize well across both common and rare descriptors.

Moreover, leveraging co-occurrence relationships could further enhance model performance by improving the understanding of odor descriptors that often appear together. This could be achieved by modeling label dependencies or using techniques that incorporate these co-occurrence patterns, leading to better predictions of multi-label odor descriptors.

\subsection{Supplementary Experiments}
\label{appendix:Supplementary Experiments}

\begin{table}
\fontsize{4pt}{5pt}\selectfont
    \centering
    \resizebox{0.3\textwidth}{!}{
    \begin{tabular}{ccr}
    \hline
\(c\) & F1 score        & AUROC           \\  \hline
0.1   & 0.4768          & 0.9316          \\
0.2   & \textbf{0.4861} & \textbf{0.9316} \\
0.3   & 0.4823          & 0.9311          \\
0.4   & 0.4847          & 0.9298          \\
1     & 0.4844          & 0.9308    \\
10    & 0.4163          & 0.9205              \\  \hline
    \end{tabular}}
    \caption{Experimental validation results of hyperparameter \(c\).}
    \label{tab:5}
\end{table}

\noindent\textbf{Hyperparametric Studies.} The hyperparameter \(c\) plays a crucial role in adjusting the energy targets for each odor descriptor in the Chemically-Informed Loss (CIL) function. It directly influences the adjustment of the energy targets \(m_{\mathrm{in}}\) and \(m_{\mathrm{out}}\), which are calculated based on the co-occurrence relationships of odor descriptors across the dataset. As shown in Table 5, the results of experiments with different values of \(c\), with the highest F1 score (0.4861) and AUROC (0.9316) achieved when \(c = 0.2\).

The value of \(c\) controls the extent to which the co-occurrence relationships between odor descriptors influence the energy targets \(m_{\mathrm{in}}\) and \(m_{\mathrm{out}}\). When \(c\) is set too low, the model may fail to adequately adjust the energy targets based on label co-occurrence, potentially leading to poor alignment with real-world odor descriptor relationships. On the other hand, when \(c\) is too high, the model becomes overly sensitive to the co-occurrence matrix, potentially overfitting to these label dependencies and losing focus on the core task of odor descriptor prediction. This leads to a drop in both F1 score and AUROC. 
 
\begin{table}
\fontsize{5pt}{6pt}\selectfont
    \centering
    \resizebox{0.48\textwidth}{!}{
    \begin{tabular}{llllll}
    \hline
\(\lambda_1\) & \(\lambda_2\)&\(\lambda_3\)& \(\lambda_4\) & F1 score         & AUROC  \\   \hline
0.5           & 0.2          & 0.2         & 0.2           & 0.4771           & 0.9231 \\
0.5           & 0.2          & 0.1         & 0.2           & 0.4805           & 0.9254 \\
1             & 0.2          & 0,1         & 0.2           & \textbf{0.4861}  & \textbf{0.9316} \\
1             & 0.2          & 0.2         & 0.2           & 0.4853           & 0.9296 \\
1             & 0.3          & 0,3         & 0.3           & 0.4837           & 0.9297 \\
1             & 0.4          & 0.4         & 0.4           & 0.4810           & 0.9240  \\  \hline
    \end{tabular}}
    \caption{Experimental validation results of hyperparameter \(\lambda_1\). \(\lambda_2\), \(\lambda_3\), and \(\lambda_4\).}
    \label{tab:Table 6}
\end{table}

As shown in Table 6, the choice of hyperparameter values \(\lambda_1 = 1\), \(\lambda_2 = 0.2\), \(\lambda_3 = 0.1\), \(\lambda_4 = 0.2\)  reflects the balance required to achieve optimal model performance, as seen in the F1 score and AUROC results. Each value is carefully selected to address specific aspects of the model's training process.

\(\lambda_1 = 1\) places the most significant emphasis on the basis loss \(\mathcal{L}_{\mathrm{basis}}\). This loss is crucial for capturing the core relationship between the molecular structure and odor descriptors. By setting \(\lambda_1\) to 1, we ensure that the model is primarily guided by the fundamental task of predicting odor descriptors, which forms the backbone of its performance. A larger value for \(\lambda_1\) means the model will focus heavily on this central component, ensuring that the foundational prediction mechanism is robust. This is essential because, without a strong basis loss, the model might fail to capture the key structural information needed for odor prediction, leading to lower accuracy; \(\lambda_2 = 0.2\) ensures the classification loss \(\mathcal{L}_{\mathrm{class}}\) contributes adequately to model accuracy, but does not dominate the learning process. The classification loss helps the model focus on predicting the correct labels for each sample. Since odor descriptor prediction is often a multi-label problem with imbalanced classes, \(\lambda_2 = 0.2\) allows the classification loss to guide the model towards improving prediction accuracy without overshadowing the other components. This moderate weight ensures that the model remains balanced, preventing overfitting to the most frequent descriptors, while still improving overall classification performance; \(\lambda_3 = 0.1\) assigns a smaller weight to the sample-level loss \(\mathcal{L}_{\mathrm{sample}}\), which helps improve consistency in individual predictions. By giving it a value of 0.1, we ensure that this component contributes to model stability without overpowering the other losses. A smaller value prevents the sample-level constraints from dominating the optimization process, allowing the model to generalize better. At the same time, it still enforces a degree of consistency across predictions, ensuring the model does not make drastic fluctuations for individual samples. This balance ensures that the model can focus on learning broader patterns rather than overfitting to specific data points; \(\lambda_4 = 0.2\) strengthens the correlation loss \(\mathcal{L}_{\mathrm{col}}\), which captures the relationships between odor descriptors. By setting this value to 0.2, we allow the model to incorporate label correlations, which are important for predicting odor descriptors that frequently occur together, such as ``fruit" and ``green". This weight strikes a balance, ensuring the model learns the co-occurrence patterns without making these correlations too dominant. Understanding label dependencies is important because it improves the model's ability to predict multiple correlated descriptors simultaneously, reflecting real-world odor experiences. A moderate weight for \(\lambda_4\) ensures that the model captures these patterns without losing focus on the individual odor descriptor predictions.

\begin{table}
\fontsize{4pt}{5pt}\selectfont
    \centering
    \resizebox{0.3\textwidth}{!}{
    \begin{tabular}{ccr}
    \hline
w/o models & F1 score        & AUROC           \\   \hline
HMFNet     & \textbf{0.4861} & \textbf{0.9316} \\
w/o HMFM   & 0.4405          & 0.9262         \\
w/o LMFE   & 0.4381          & 0.9227         \\
w/o GMFE   & 0.4757          & 0.9233              \\  \hline
    \end{tabular}}
    \caption{Ablation studies were conducted on structural variants of HMFNet.}
    \label{tab7}
\end{table}

\noindent\textbf{Ablation Studies.} We evaluate the performance of HMFNet and its variants by systematically removing specific components. Specifically, w/o HMFM refers to the removal of the Harmonic Modulated Feature Mapping, w/o LMFE refers to the removal of the fine-grained Local Multi-Hierarchy Feature Extraction Module, and w/o GMFE refers to the removal of the Global Multi-Hierarchy Feature Extraction Module. 

Table 7 presents the ablation study results comparing these variants, highlighting the following key findings: Removing the Harmonic Modulated Feature Mapping (HMFM) module results in a performance drop, with an F1 score of 0.4405 and AUROC of 0.9262, demonstrating that HMFM contributes to capturing complex molecular-odor relationships. Similarly, excluding the fine-grained Local Multi-Hierarchy Feature Extraction Module (LMFE) leads to a further decrease, with an F1 score of 0.4381 and AUROC of 0.9227, indicating that fine-grained local features are essential for improving model accuracy. Removing the Global Multi-Hierarchy Feature Extraction Module (GMFE) also diminishes performance, with an F1 score of 0.4757 and AUROC of 0.9233, emphasizing the importance of capturing global molecular structures for long-range dependencies. These findings underscore the significance of each component in HMFNet. The full model achieves the best performance, highlighting the synergy between fine-grained local feature extraction, global structure understanding, and harmonic modulation of features.

\begin{figure*}
\centerline{\includegraphics[width=1\textwidth,keepaspectratio]{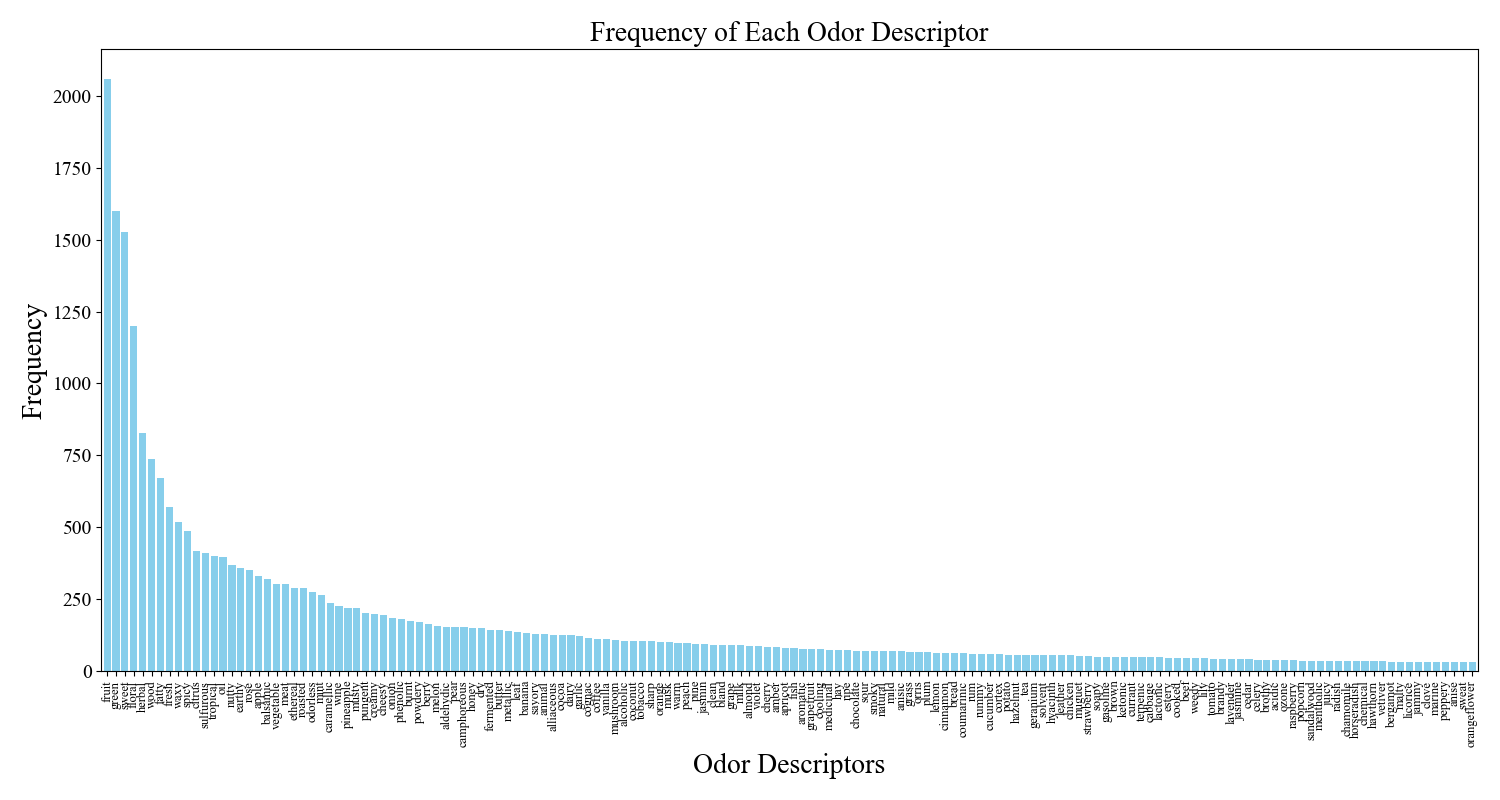}}
\caption{Distribution of odor descriptor frequency in dataset.}
\label{fig}
\end{figure*}

\begin{figure*}
\centerline{\includegraphics[width=0.7\textwidth,keepaspectratio]{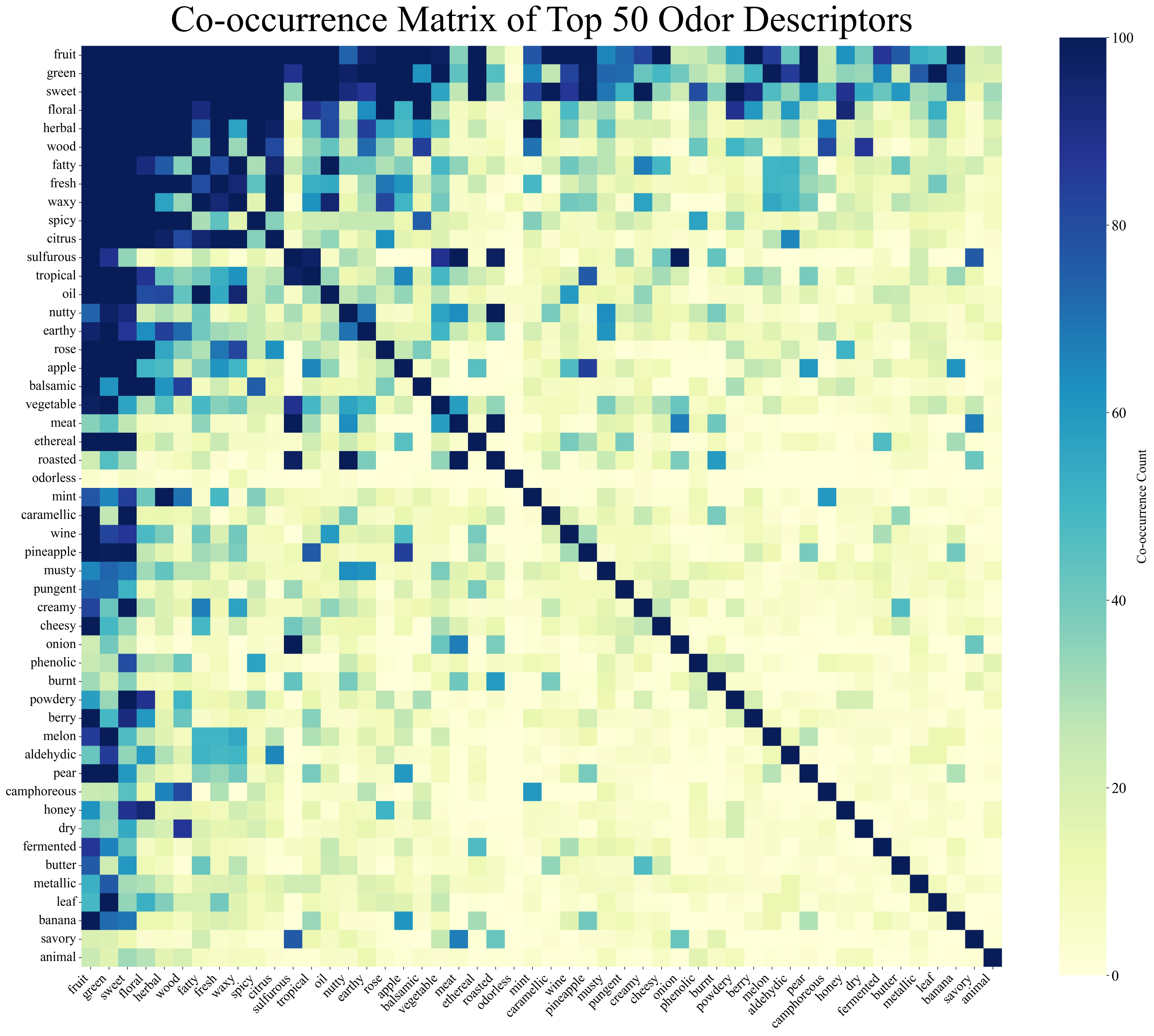}}
\caption{Co-ocurrence matrix for odor descriptors.}
\label{fig}
\end{figure*}

\end{document}